\documentclass[runningheads]{llncs}
\usepackage{graphicx}

\usepackage{tikz}
\usepackage{comment}
\usepackage{amsmath,amssymb} 
\usepackage{color}

\usepackage[accsupp]{axessibility}  

\usepackage{enumitem}
\usepackage{booktabs}
\usepackage{makecell}
\usepackage{multirow} 
\usepackage{xcolor}
\usepackage{colortbl}
\definecolor{mygray}{gray}{0.6}
\definecolor{background_gray}{gray}{0.9}

\usepackage{floatrow}
\newfloatcommand{capbtabbox}{table}[][\FBwidth]

\usepackage{hyperref}
\newlength\savedwidth

\newlength\savewidth
\newcommand\shline{\noalign{\global\savewidth\arrayrulewidth
                            \global\arrayrulewidth 1.5pt}%
                   \hline
                   \noalign{\global\arrayrulewidth\savewidth}}

\begin{document}

\pagestyle{headings}
\mainmatter
\def\ECCVSubNumber{1596}  

\title{DualFormer: Local-Global Stratified Transformer for Efficient Video Recognition} 

\titlerunning{Local-Global Stratified Transformer for Efficient Video Recognition}
%
\author{Yuxuan Liang\inst{1,2} \and
Pan Zhou\inst{1} \and 
Roger Zimmermann\inst{2} \and
Shuicheng Yan\inst{1}}
\authorrunning{Liang et al.}
\institute{$^1$ Sea AI Lab $^2$ National University of Singapore \\
\email{\{yuxliang,rogerz\}@comp.nus.edu.sg \{zhoupan,ysc\}@sea.com}} 
\maketitle

\begin{abstract}
While transformers have shown great potential on video recognition with their strong capability of capturing long-range dependencies, they often suffer high computational costs induced by the self-attention to the huge number of 3D tokens. In this paper, we present a new transformer architecture termed DualFormer, which can efficiently perform space-time attention for video recognition. Concretely, DualFormer stratifies the full space-time attention into dual cascaded levels, i.e., to first learn fine-grained local interactions among nearby 3D tokens, and then to capture coarse-grained global dependencies between the query token and global pyramid contexts. Different from existing methods that apply space-time factorization or restrict attention computations within local windows for improving efficiency, our local-global stratification strategy can well capture both short- and long-range spatiotemporal dependencies, and meanwhile greatly reduces the number of keys and values in attention computation to boost efficiency. Experimental results verify the superiority of DualFormer on five video benchmarks against existing methods. In particular, DualFormer achieves 82.9\%/85.2\% top-1 accuracy on Kinetics-400/600 with $\sim$1000G inference FLOPs which is at least 3.2$\times$ fewer than existing methods with similar performance. We have released the source code at \url{https://github.com/sail-sg/dualformer}.

\end{abstract}

\section{Introduction} \label{sec:intro}
Video recognition is a fundamental task in computer vision, such as action recognition~\cite{carreira2017i3d} and event detection~\cite{hongeng2004video}.
Like in image-based tasks~\cite{krizhevsky2012imagenet,simonyan2014very,he2016deep}, Convolutional Neural Networks (CNNs) are often taken as backbones for video recognition models~\cite{lin2019tsm,du2015c3d,tran2018r21d,carreira2017i3d,feichtenhofer2020x3d,feichtenhofer2019slowfast}.
Though successful, it is challenging for convolutional architectures to capture \textit{long-range spatiotemporal dependencies} across video frames due to their limited receptive field.

Recently, transformers~\cite{vaswani2017attention} have become an alternative paradigm for visual modeling beyond CNNs, demonstrating great potential in a series of image processing tasks~\cite{wang2021pyramid,liu2021Swin,ranftl2021vision,wang2021end,wang2021max,zhu2020deformable}. 
A pioneering work is the Vision Transformer (ViT)~\cite{dosovitskiy2020image} which replaces the inherent inductive bias of locality in convolutions by global relation modeling with multi-head self-attention (MSA)~\cite{vaswani2017attention}.
Soon the vision community extends the application of MSA from static images to videos considering its remarkable power for capturing long-range spatiotemporal dependencies~\cite{neimark2021video,fan2021multiscale,bertasius2021space,arnab2021vivit}.
Concretely, a video is first partitioned into non-overlapping 3D patches, similar as in NLP tasks~\cite{vaswani2017attention}, which then serve as input tokens for transformers to jointly learn short- and long-range relations within a video. 
\begin{figure}[!t]
  \centering
  \includegraphics[width=0.87\textwidth]{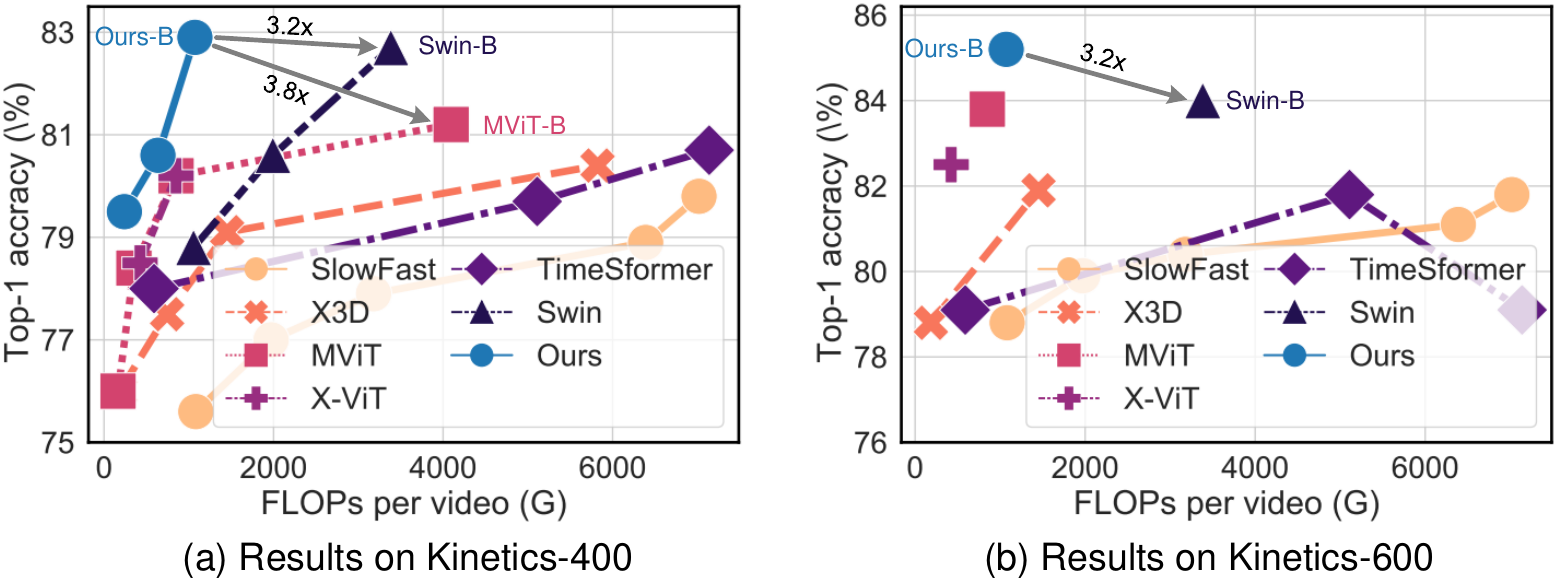}
  \caption{Accuracy vs. FLOPs on Kinetics~\cite{kay2017kinetics}. Ours-B is the base version of DualFormer.}
  \label{fig_acc_flops}
\end{figure}

One of the major challenges for applying transformers to video data is their \textit{low efficiency}.
Due to the MSA operation, the computational cost of video transformers grows quadratically with the increasing number of tokens, and may even become totally unaffordable for some high spatial resolution or long videos. 
To alleviate this issue, TimeSformer~\cite{bertasius2021space} and ViViT~\cite{arnab2021vivit} factorize the full space-time self-attention along temporal and spatial dimensions separately to achieve a  balance between accuracy and efficiency in video recognition. Inspired by the observation that near tokens are usually more related than distant ones~\cite{tobler1970computer}, Video Swin Transformer~\cite{liu2021video} applies the inductive bias of locality at each transformer layer via performing  self-attention in the non-overlapping local windows. 
Though effective, both the space-time factorization and the local-window based attention scheme contradict the aim of applying full space-time attention, i.e., to \emph{jointly} capture local and global spatiotemporal dependencies within one layer, and thus impair  the performance of video transformers. 

In this work, we present a new video transformer architecture entitled \textbf{DualFormer} for \emph{efficient} video recognition. DualFormer stratifies the full space-time attention into dual cascaded levels: 1) \emph{Local-Window based Multi-head Self-Attention} (LW-MSA) to  extract short-range interactions among nearby tokens; and 2) \emph{Global-Pyramid based MSA} (GP-MSA) to  capture long-range dependencies between the query token and the coarse-grained global pyramid contexts. In this manner, DualFormer significantly reduces the number of keys and values in attention computation, and achieves much higher efficiency over existing video transformers \cite{bertasius2021space,arnab2021vivit,liu2021video} with comparable performance, as shown in Figure \ref{fig_acc_flops}.

\begin{figure}[!t]
  \centering
  \includegraphics[width=0.98\textwidth]{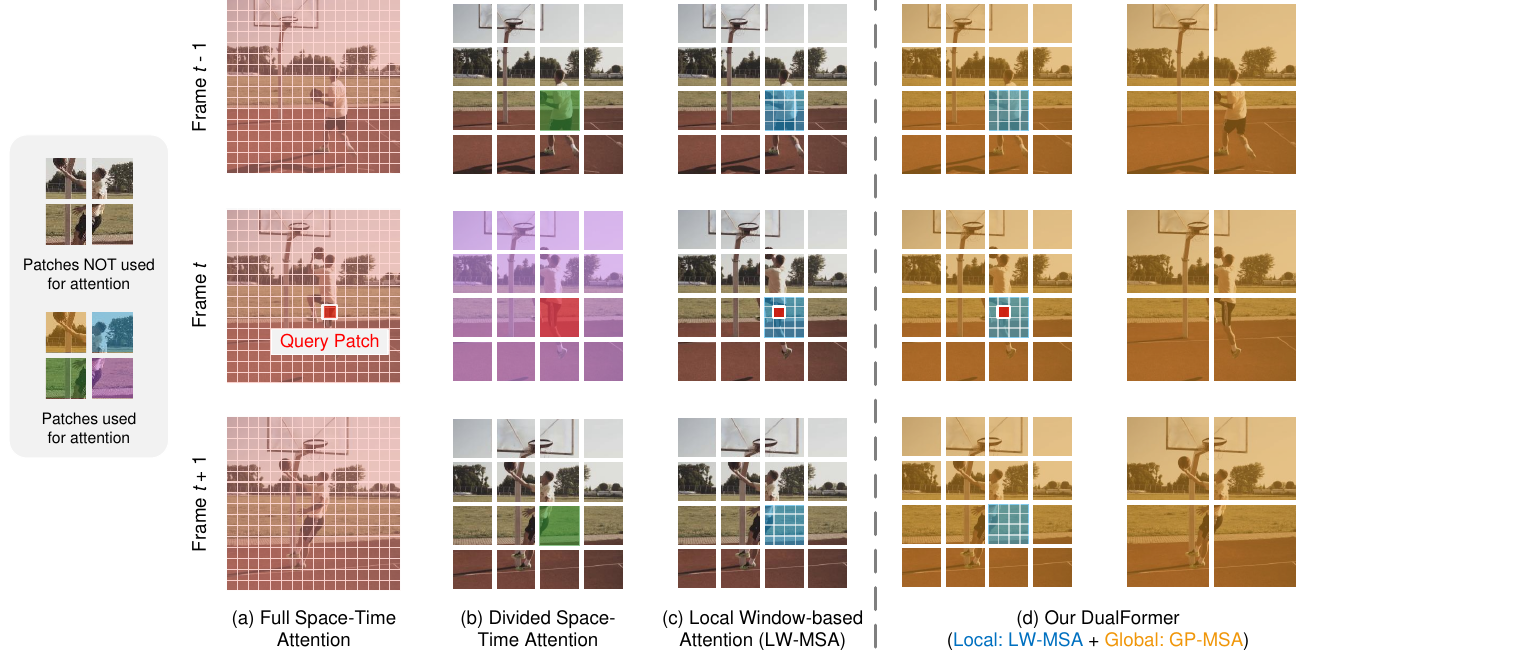}
  \caption{Visualization of four space-time MSA schemes. For better illustration, we use 2D patch partitions. We denote in red the query patch and in non-red colors its attention targets for each scheme.
  Multiple highlighted colors in a scheme indicate the MSA separately applied along different dimensions. 
  (a) Full space-time attention~\protect\cite{bertasius2021space} has quadratic complexity w.r.t. the number of patches. 
  (b) Divided space-time attention~\protect\cite{bertasius2021space}, where MSA is separately applied in temporal and spatial domains. 
  (c) Local window-based attention~\protect\cite{liu2021video} improves efficiency by restricting MSA computation within local windows, lacking interactions between distant patches.
  (d) Our dual-level MSA scheme stratifies the modeling of local and global relations. Given a query patch, we first use LW-MSA to compute attention weights within the local window. Then, the query patch attends to the multi-scale global priors (two scales here) via GP-MSA.}
  \label{fig_intro}
\end{figure}

Figure \ref{fig_intro} shows how a query patch (in red) attends to its surroundings in a DualFormer block.
Following the intuition that tokens closer to each other are more likely to be correlated~\cite{liu2021Swin,yang2021focal},
we first perform LW-MSA at a fine-grained level to allow each patch to interact with its neighbors within a local window. 
This strategy has also been verified to be efficient and memory-friendly by recent studies~\cite{liu2021Swin,yang2021focal,chu2021Twins,liu2021video,chen2021regionvit}.
Next, at the global level, a query patch attends to the full region of interest at a coarse granularity via GP-MSA. 
To be specific, we first extract global contextual priors with different pyramid scales for multi-scale scene interpretation (see the two scales, i.e. small windows and large windows, in Figure \ref{fig_intro}(d)).
These global priors then pass global contextual information to the query tokens via MSA. 
Since such priors are extracted at a coarse-grained level, their number is much smaller than the original token number, leading to far less computation cost in capturing global information than the full space-time attention. In contrast to the space-time factorization in TimeSformer \cite{bertasius2021space} and ViViT \cite{arnab2021vivit} and the locality-based scheme in Swin \cite{liu2021video}, this dual-level attention design not only enables our model to have the global receptive field at each block, but is also efficient in attention computations. 

Extensive experimental results on five video benchmarks  validate the superiority of our DualFormer in terms of accuracy and FLOPs. In particular, our DualFormer achieves 82.9\%/85.2\% top-1 accuracy on Kinetics-400/600~\cite{kay2017kinetics} with only $\sim$1000 GFLOPs which is 3.2$\times$ and 16.2$\times$ fewer than the previous state-of-the-art methods, i.e., Swin~\cite{liu2021Swin} and ViViT~\cite{arnab2021vivit}, respectively. We strongly believe that such gains on efficiency benefit real-world deployments of video recognition models, especially for deployments on edge devices. See detailed comparison  on Kinetics-400/600 in Figure \ref{fig_acc_flops}. Furthermore, our model also achieves state-of-the-art performance on three smaller datasets under transfer learning settings.

\section{Related Work}
\textbf{CNNs for Video Recognition.} 
CNN-based video recognition models can be categorized into two groups: 2D CNNs and 3D CNNs~\cite{li2021survey}.
For the first group~\cite{karpathy2014large,wang2018temporal}, each video frame is processed separately by 2D convolutions and then aggregated along the time axis at the top of the network.
However, some studies point out that 2D convolutions cannot well capture the information along the temporal dimension~\cite{lin2019tsm,liu2021tam,liu2020teinet,jiang2019stm}. 
The second group learns spatiotemporal video representation via 3D convolutions by aggregating space-time features and are difficult to optimize~\cite{ji20123d,du2015c3d,xu2017r,hara2018can,hara2017learning}.
Thus, the current trend for 3D CNN-based video recognition is to boost efficiency.
For example, I3D~\cite{carreira2017i3d} expands pretrained 2D CNNs~\cite{krizhevsky2012imagenet,simonyan2014very,he2016deep} into 3D CNNs;
some recent works~\cite{qiu2017p3d,xie2018s3d,tran2018r21d,tran2019video,feichtenhofer2019slowfast,feichtenhofer2020x3d} factorize 3D convolutions into spatial and temporal filters, demonstrating even higher accuracy than vanilla 3D CNNs.
Unfortunately, most of the 2D and 3D CNNs cannot capture long-range spatiotemporal dependencies due to their limited receptive fields, which leads to sub-optimal recognition performance. 

\noindent \textbf{Transformers for Video Recognition.}\label{para:related}
Recently, transformers are applied to model spatiotemporal dependencies for video recognition~\cite{neimark2021video,fan2021multiscale,bertasius2021space,bulat2021space,arnab2021vivit,zhang2021vidtr,liu2021video,patrick2021keeping} by virtue of their great power in capturing long-range dependencies~\cite{dosovitskiy2020image,touvron2021training,liu2021Swin,chu2021Twins}. 
With pretraining on a large-scale image dataset, video transformers achieve promising  performance on video benchmarks~\cite{bertasius2021space,arnab2021vivit,liu2021video}, such as Kinetics-400/600. However, the potential of video transformers is significantly limited by the considerable computational complexity of performing full space-time attention. Various approaches have been proposed to reduce such computation cost~\cite{bertasius2021space,bulat2021space,arnab2021vivit,zhang2021vidtr,liu2021video}. For instance, TimeSformer~\cite{bertasius2021space} factorizes the full space-time attention into spatial and temporal dimensions. Similarly, ViViT~\cite{arnab2021vivit} examines three variants of space-time factorization for computation reduction. X-ViT~\cite{bulat2021space} approximates the space-time attention by restricting the temporal attention to a local temporal window and using a mixing strategy. Video Swin Transformer~\cite{liu2021video} introduces an inductive bias of locality to transformers for video understanding. However, these attempts focus on either space-time factorization or restricting attention computation locally, crippling the capability of MSA in capturing long-range dependencies. To solve this, we present a new transformer called DualFormer to improve the efficiency of video transformers, by alternatively capturing fine-grained local interactions and coarse-grained global information within each block. Besides, there are some concurrent works enhancing transformers \cite{zha2021shifted,li2021uniformer} or exploring self-supervised pretraining schemes \cite{qian2021spatiotemporal,wei2021masked} for video recognition.

\section{Methodology}
We start by summarizing the overall architecture of DualFormer in Sec.~\ref{para:framework}, and then elaborate on its basic block in Sec.~\ref{para:block} by well introducing the two types of attention, including Local-Window based Multi-head Self-Attention (LW-MSA) and Global-Pyramid based MSA (GP-MSA). Afterward, we explain the network configuration for constructing our DualFormer in Sec.~\ref{para:config}. Finally, we discuss the differences between our DualFormer and related works in Sec.~\ref{para:discussion}.

\subsection{Overall Architecture}\label{para:framework}
\begin{figure}[!t]
  \centering
  \includegraphics[width=0.99\textwidth]{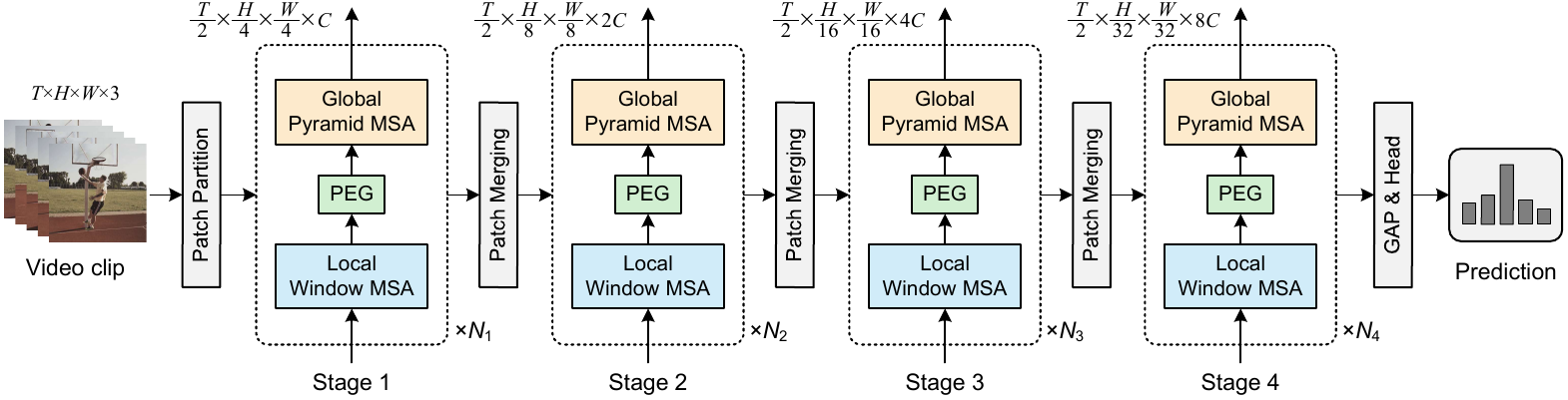}
  \caption{Overall architecture of DualFormer. GAP: global average pooling.}
  \label{fig_framework}
\end{figure}

Figure \ref{fig_framework} shows the overall architecture of the proposed DualFormer.
It takes a video clip $\mathcal{X} \in \mathbb{R}^{T \times H \times W \times 3}$ as input, where $T$ stands for the number of frames and each frame consists of $H \times W \times 3$ pixels. 
To accommodate  high-resolution video-based tasks, our model leverages a hierarchical design~\cite{fan2021multiscale,yang2021focal,liu2021Swin,liu2021video} to produce decreasing-resolution feature maps from early to late stages.
First, we partition a video clip into non-overlapping 3D patches of size $2 \times 4 \times 4 \times 3$ and employ a linear layer for projection, resulting in $\frac{T}{2} \times \frac{H}{4} \times \frac{W}{4}$ visual tokens with feature channel dimension $C$. Then, as shown in Figure \ref{fig_framework},  these tokens  go through the four stages of DualFormer for learning visual representations. 
At each stage $s \in \{1,2,3,4\}$, we sequentially stack $N_s$ DualFormer blocks for spatiotemporal learning, where $N_s$ controls the capacity of each model stage.
Each DualFormer block consists of dual cascaded levels of self-attention mechanisms: LW-MSA for learning short-range interactions within local windows, and GP-MSA for capturing long-range context information within the whole video. Additionally, a convolution-based Position Encoding Generator (PEG) \cite{chu2021we} is integrated into the first block of each stage (between the two types of MSA) to empower position-aware self-attention. After each stage, DualFormer follows the prior art~\cite{liu2021video} to utilize a patch merging layer to downsample the spatial size of the feature map by 2$\times$, while the feature channel dimension is increased by 2$\times$. Once the output of the last stage is obtained, DualFormer performs video recognition by applying a global average pooling (GAP) layer followed by a linear classifier. 

\subsection{DualFormer Block}\label{para:block}
As all blocks share the same architecture,  we introduce each block by taking a block at the $s$-th stage as an example. Assume that the input feature map at the $s$-th stage is of resolution $T_s \times H_s  \times W_s$  with channel dimension $C_s$, the complexity of the full space-time attention is $\mathcal{O}(T_s^2 H_s^2 W_s^2 C_s)$ which is too high to handle high-resolution videos in practice. To alleviate this efficiency issue, 
in each DualFormer block, we stratify the full space-time attention into dual cascaded levels,  i.e., to first learn fine-grained local space-time interactions among nearby 3D tokens by our LW-MSA, and then to capture  coarse-grained global dependencies between the query token and the coarse-grained global pyramid contexts via our GP-MSA. Next, we will delineate LW-MSA and GP-MSA. 

\subsubsection{Local-Window based MSA.} 
Considering nearby tokens often have stronger correlations than faraway tokens, we perform LW-MSA to compute the self-attention within non-overlapping 3D windows to capture local interactions among tokens.
As shown in  Figure \ref{fig_intro}{\color{red}d}, given a feature map with $T_s\times H_s \times W_s$ patch tokens with dimension $C_s$,  we first evenly split it into non-overlapping small local windows, each of which is of size $t_s \times h_s \times w_s$, yielding $\frac{T_s}{t_s} \times \frac{H_s}{h_s} \times \frac{W_s}{w_s}$ windows. Next, we flatten all tokens within the $(i,j,k)$-th local window into $\mathbf{X}_{i,j,k} \in \mathbb{R}^{t_sh_sw_s \times C_s}$. Now we are ready to formulate our LW-MSA: 
\begin{equation}\label{eq:msa}
	\setlength{\abovedisplayskip}{4pt}
	\setlength{\belowdisplayskip}{4pt}
	\setlength{\abovedisplayshortskip}{4pt}
	\setlength{\belowdisplayshortskip}{4pt}
\mathbf{X}^{\prime}_{ijk}  =\operatorname{MSA}(\operatorname{LN}(\mathbf{X}_{ijk})) + \mathbf{X}_{ijk}, \quad 
\mathbf{Y}_{ijk}  = \operatorname{MLP}\left(\operatorname{LN}\left(\mathbf{X}^{\prime}_{ijk}\right)\right)+\mathbf{X}^{\prime}_{ijk},
\end{equation}
where $\operatorname{MSA}$, $\operatorname{LN}$, and $\operatorname{MLP}$ denote a standard multi-head self-attention, a  layer normalization~\cite{ba2016layer}, and a multi-layer perceptron, respectively. The computational complexity\footnote{For simplicity, we omit the complexity of MLP in this paper.} of MSA within a local window is computed as $\mathcal{O}((t_sh_sw_s)^2 C_s)$. We further summarize the cost of all $\frac{T_s}{t_s} \times \frac{H_s}{h_s} \times \frac{W_s}{w_s}$ windows as follows: 
\begin{equation}\label{eq:o_lw_msa}
	\setlength{\abovedisplayskip}{4pt}
	\setlength{\belowdisplayskip}{4pt}
	\setlength{\abovedisplayshortskip}{4pt}
	\setlength{\belowdisplayshortskip}{4pt}
\mathcal{O}(\operatorname{LW-MSA}) = (t_sh_sw_s)^2 C_s \times \left(\frac{T_s H_s W_s}{t_s h_s w_s}\right) = t_s h_s w_s M_s C_s, 
\end{equation}
where $M_s=T_s H_s W_s $ is the token number. 
In this way, the complexity of our LW-MSA is $\frac{M_s}{t_sh_sw_s} \times$ less than that of full space-time attention $\mathcal{O}(M_s^2 C_s)$. Since videos often have a huge number of tokens ($M_s$ is large) and the local window is of small size ($t_s h_s w_s$ is small), our LW-MSA enjoys much higher efficiency for video recognition. See the effects of $t_s,h_s,w_s$ on the performance in Sec. \ref{para:ablation}.

\subsubsection{Global-Pyramid based MSA.}\label{para:gpmsa} 
While being efficient in computation, LW-MSA cripples the ability of MSA to capture global information. 
For example, a query patch cannot attend to a patch outside the local window. 
To tackle this issue, a shifted window strategy is proposed to enable a patch to communicate with the patches inside adjacent windows in~\cite{liu2021video}. 
Nevertheless, it is still difficult for patches to interact with distant windows. 
In this work, we propose GP-MSA as a complement for learning long-range dependencies within the whole video. 

\begin{figure}[!t]
  \centering
  \includegraphics[width=0.78\textwidth]{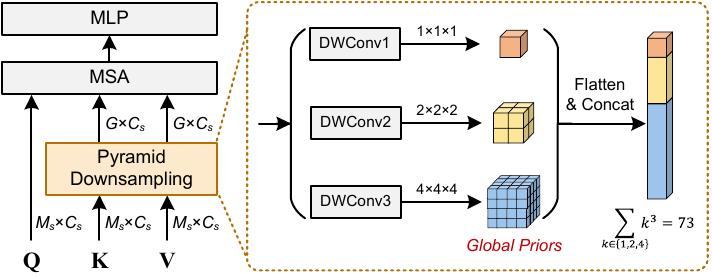}
  \caption{The pipeline of GP-MSA. DWConv denotes depth-wise convolution~\protect\cite{chollet2017xception} for generating global priors with multiple scales. For simplicity, we use a three-level pyramid (1$\times$1$\times$1, 2$\times$2$\times$2, and 4$\times$4$\times$4) for illustration.  With pyramid downsampling, the computational cost and memory usage of GP-MSA are much lower than those of standard MSA due to reduction of the key/value number. }
  \label{fig_gp_msa}
\end{figure}

As a variant of MSA, our GP-MSA receives queries $\mathbf{Q}$, keys $\mathbf{K}$, and values $\mathbf{V}$ as input to capture the global information. 
For simplicity, we assume $\mathbf{Q}$, $\mathbf{K}$ and $\mathbf{V}$ are all in shape $M_s \times C_s$, where $M_s$ is the number of tokens at the $s$-th stage. 
Different from the vanilla MSA, our GP-MSA proposes a simple yet effective method, termed \textbf{pyramid downsampling}, to reduce the spatiotemporal scale of $\mathbf{K}$ and $\mathbf{V}$ before performing MSA, so as to lessen the computational overheads and memory usage.  
Specifically, as illustrated by Figure \ref{fig_gp_msa}, our pyramid downsampling adopts three levels of depth-wise convolutions~\cite{chollet2017xception} to generate a set of global priors, where each prior is a spatiotemporal abstract of the original feature map under different pyramid scales. 
This operation allows the model to separate the feature map into non-overlapping regions and to build pooled representations for various locations.  
For example, the 1$\times$1$\times$1 prior (the orange cube in Figure \ref{fig_gp_msa}) denotes the coarsest scale with only a single value at each channel, which is similar to global average pooling~\cite{lin2014network} that covers the whole video, while the 2$\times$2$\times$2 prior (the yellow cube) indicates a summary of finer granularity. 
Then, as shown in Figure \ref{fig_gp_msa}, we flatten and concatenate these priors to be the new key-value of size $G \times C_s$, where $G$ denotes the number of space-time locations, e.g., $G$=$\sum_{k=\{1,2,4\}}{k^3}$=73. After downsampling, we pass the global contextual information in these priors to each query patch via standard MSA. 

\noindent \emph{Complexity Analysis of GP-MSA.}
Without loss of generality, assume we have $N_g$ pyramid scales for all stages and denote the size of global prior at the $i$-th scale as $(k_1^i, k_2^i, k_3^i)$, where $k_1^i < T_s$, $k_2^i < H_s$ and $k_3^i < W_s$. The complexity of GP-MSA at the $s$-th stage is computed as:
\begin{equation*}
	\setlength{\abovedisplayskip}{4pt}
	\setlength{\belowdisplayskip}{3.5pt}
	\setlength{\abovedisplayshortskip}{4pt}
	\setlength{\belowdisplayshortskip}{3.5pt}
	\small
\begin{split}
\mathcal{O}(\operatorname{GP-MSA}) &= \underbrace{M_s C_s \sum_{i=1}^{N_g}{k^i_1 k^i_2 k^i_3}}_{\operatorname{MSA}} + \underbrace{\sum_{i=1}^{N_g}\left( \frac{T_s H_s W_s}{k^i_1 k^i_2 k^i_3}k^i_1 k^i_2 k^i_3 C_s\right)}_{\operatorname{DWConv}} \\
&= \bigg(\sum_{i=1}^{N_g}{k^i_1 k^i_2 k^i_3} + N_g \bigg)M_s C_s=(G+N_g) M_s C_s, \\
\end{split}
\end{equation*}
where $G=\sum_{i=1}^{N_g}{k^i_1 k^i_2 k^i_3}$ is the number of global priors, i.e., new keys or values after reduction. To further improve efficiency, we draw inspiration from R(2+1)D~\cite{tran2018r21d} to factorize the depth-wise convolution at each scale along temporal and spatial dimensions, which gives an   even less complexity:
\begin{equation}\label{eq:gp_msa}
	\setlength{\abovedisplayskip}{4pt}
	\setlength{\belowdisplayskip}{4pt}
	\setlength{\abovedisplayshortskip}{4pt}
	\setlength{\belowdisplayshortskip}{4pt}
	\small 
\mathcal{O}(\operatorname{GP-MSA})  =\! \bigg(G+\sum_{i=1}^{N_g}\!\left(\frac{k_1^i}{T_s} + \frac{k_2^i k_3^i}{H_s W_s}\right)\!\bigg)\! M_s C_s  
 \approx G M_s C_s < \underbrace{(G+N_g) M_s C_s}_{\text{Previous}}\ll \underbrace{M_s^2 C_s}_{\operatorname{MSA}}.
\end{equation}
Since $G$ is generally much smaller than the number of tokens ($M_s$) in the original feature map, our GP-MSA significantly reduces the computational complexity and memory usage during learning global representations. For instance, at the first stage of DualFormer where $G$ is 456 while $M_s$ is 50176, the complexity has been reduced by $\sim$110 times.
In a nutshell, the overall complexity of MSA in a DualFormer block is the summation of LW-MSA in Eq. (\ref{eq:o_lw_msa}) and GP-LSA in Eq.~(\ref{eq:gp_msa}), and is much smaller than the complexity of vanilla full-time self-attention, demonstrating the efficiency of our DualFormer.

\begin{table}[!b]
  \centering
  \scriptsize
  \tabcolsep=3mm
    \begin{tabular}{c|c|c|c|c}
    \shline
    Stage & Layer & \textbf{Tiny} & \textbf{Small} & \textbf{Base} \\
    \hline
    \multirow{2}{*}{\makecell{Stage 1 \\ \\ Output: \\ $\frac{T}{2},\frac{H}{4},\frac{W}{4}$}} & \makecell{Patch\\Merging} & \makecell{$p_1=(2,4,4)$\\$C_1=64$}    & \makecell{$p_1=(2,4,4)$\\$C_1=96$}    & \makecell{$p_1=(2,4,4)$\\$C_1=128$} \\
\cline{2-5}            & \makecell{\color{blue}{LW-MSA}\\\color{brown}{GP-MSA}} &  $\left[   \begin{array}{l} \color{blue}{(8,7,7)} \\ \color{brown}{(4,4,4)} \\ \color{brown}{(8,7,7)} \end{array} \right]{\times 1}$     &  $\left[   \begin{array}{l} \color{blue}{(8,7,7)} \\ \color{brown}{(4,4,4)} \\ \color{brown}{(8,7,7)} \end{array} \right]{\times 1}$     & $\left[   \begin{array}{l} \color{blue}{(8,7,7)} \\ \color{brown}{(4,4,4)} \\ \color{brown}{(8,7,7)} \end{array} \right]{\times 1}$ \\
    \hline
\multirow{2}{*}{\makecell{Stage 2 \\ \\ Output: \\ $\frac{T}{2},\frac{H}{8},\frac{W}{8}$}} & \makecell{Patch\\Merging} & \makecell{$p_2=(1,2,2)$\\$C_2=128$}    & \makecell{$p_2=(1,2,2)$\\$C_2=192$}    & \makecell{$p_2=(1,2,2)$\\$C_2=256$} \\
\cline{2-5}            & \makecell{\color{blue}{LW-MSA}\\\color{brown}{GP-MSA}} &  $\left[   \begin{array}{l} \color{blue}{(8,7,7)} \\ \color{brown}{(4,4,4)} \\ \color{brown}{(8,7,7)} \end{array} \right]{\times 1}$     &  $\left[   \begin{array}{l} \color{blue}{(8,7,7)} \\ \color{brown}{(4,4,4)} \\ \color{brown}{(8,7,7)} \end{array} \right]{\times 1}$     & $\left[   \begin{array}{l} \color{blue}{(8,7,7)} \\ \color{brown}{(4,4,4)} \\ \color{brown}{(8,7,7)} \end{array} \right]{\times 1}$ \\
    \hline
\multirow{2}{*}{\makecell{Stage 3 \\ \\ Output: \\ $\frac{T}{2},\frac{H}{16},\frac{W}{16}$}} & \makecell{Patch\\Merging} & \makecell{$p_3=(1,2,2)$\\$C_3=256$}    & \makecell{$p_3=(1,2,2)$\\$C_3=384$}    & \makecell{$p_3=(1,2,2)$\\$C_3=512$} \\
\cline{2-5}            & \makecell{\color{blue}{LW-MSA}\\\color{brown}{GP-MSA}} &  $\left[   \begin{array}{l} \color{blue}{(8,7,7)} \\ \color{brown}{(4,4,4)} \\ \color{brown}{(8,7,7)} \end{array} \right]{\times 5}$     &  $\left[   \begin{array}{l} \color{blue}{(8,7,7)} \\ \color{brown}{(4,4,4)} \\ \color{brown}{(8,7,7)} \end{array} \right]{\times 9}$     & $\left[   \begin{array}{l} \color{blue}{(8,7,7)} \\ \color{brown}{(4,4,4)} \\ \color{brown}{(8,7,7)} \end{array} \right]{\times 9}$ \\
    \hline
\multirow{2}{*}{\makecell{Stage 4 \\ Output: \\ $\frac{T}{2},\frac{H}{32},\frac{W}{32}$}} & \makecell{Patch\\Merging} & \makecell{$p_4=(1,2,2)$\\$C_4=512$}    & \makecell{$p_4=(1,2,2)$\\$C_4=768$}    & \makecell{$p_4=(1,2,2)$\\$C_4=1024$} \\
\cline{2-5}            & \makecell{\color{blue}{LW-MSA}\\\color{brown}{GP-MSA}} &  $\left[   \begin{array}{l} \color{blue}{(8,7,7)} \\ \color{brown}{(8,7,7)} \end{array} \right]{\times 2}$     &  $\left[   \begin{array}{l} \color{blue}{(8,7,7)} \\ \color{brown}{(8,7,7)} \end{array} \right]{\times 1}$     & $\left[   \begin{array}{l} \color{blue}{(8,7,7)} \\ \color{brown}{(8,7,7)} \end{array} \right]{\times 1}$ \\
    \shline
    \end{tabular}
  \caption{Model configurations of DualFormer, including three versions. $p_i$ and $C_i$ denote patch size and feature dimension at the $i$-th stage, respectively.}\label{tab:config}%
\end{table}%

\subsubsection{Position Encoding Generator (PEG).} 
As the self-attention operation is permutation-invariant, we draw inspiration from conditional positional encoding \cite{chu2021we} to utilize a convolution layer as a position encoding generator (PEG) to encode the position information into self-attention as follows:
\begin{equation}
\small
\operatorname{PEG}(\mathbf{X})=\operatorname{DWConv}(\mathbf{X})+\mathbf{X},
\end{equation}
where $\mathbf{X}$ is the input of the current stage. $\operatorname{DWConv}(\cdot)$ represents 3D depth-wise convolution for improving efficiency (compared with standard convolutions). Primarily, convolutions can provide \textit{absolute position} information, which has been verified in \cite{islam2020much,chu2021we}. By using convolutions, the position embedding is no longer input-agnostic and dynamically generated based on the local neighbors of each token. Moreover, our PEG is permutation-variant since the permutation over inputs affects the order in local windows. In addition, the convolution kernels are applied to local windows everywhere in an input video, thus having similar responses to the objects with similar features, i.e., translation-invariant. 

\subsection{Model Configuration}\label{para:config}
Following Swin~\cite{liu2021video}, we consider three network configurations (i.e., base, small and tiny) for our DualFormer. 
For the LW-MSA of all versions, its local window size is always $(8,7,7)$, and its MLP expansion factor is always 4. In GP-MSA, we utilize two pyramid scales $(8,7,7)$ and $(4,4,4)$ at the first three stages for learning global contextual information. At the last stage, since the   feature map size has become $(16,7,7)$, we only extract one scale of global prior with $(8,7,7)$ using a depth-wise convolution. More details can be found in Table \ref{tab:config}.

\subsection{Discussion}\label{para:discussion}
Here, we compare our model with some related works mentioned in Sec. \ref{para:related}. 

\noindent\textbf{Comparison with Space-Time Factorization.} The space-time attention factorization in TimeSformer~\cite{bertasius2021space} and ViViT~\cite{arnab2021vivit} separately perform standard MSA in temporal and spatial domains, while DualFormer has two major differences. Firstly, our DualFormer factorizes the full space-time attention along another two dimensions, namely, local and global dependencies via LW-MSA and GP-MSA respectively in which both model temporal and spatial domains as a whole and thus better capture their complementary information. Secondly, for each  domain, TimeSformer and ViViT still perform conventional MSA attention among all tokens. Differently, our LW-MSA and GP-MSA first considers the attention among nearby 3D tokens and then integrate the global information at the local-window level, which greatly reduces the number of keys and values for attention computation and boosts efficiency.

\noindent\textbf{Comparison with Video Swin.} Our DualFormer also distinguishes Swin~\cite{liu2021video} from their different ways for long-range relation modeling. In Swin, a shifting window strategy is proposed to empower cross-local-window interaction, and thus increases the receptive fields of MSA. Nevertheless, it is still non-trivial for this shifting scheme to learn the dependencies between distant patch tokens. In contrast, our DualFormer employs GP-MSA to \emph{directly} capture the interaction between the query token and the coarse-grained global pyramid contexts, which is more explicit and efficient to learn the global spatiotemporal dependencies. Experimental results in Figure \ref{fig_acc_flops} verify that DualFormer can achieve slightly higher accuracy while having at least 3$\times$ fewer FLOPs than Swin.

\noindent\textbf{Comparison with Image-based ViTs.}
Several image-based transformers with a local-to-global design, e.g.,  Twins~\cite{chu2021Twins} and RegionViT~\cite{chen2021regionvit}, are also relevant to our model. Compared to Twins, the major difference is the construction of global contexts. Since the objects across different frames in a video may vary in sizes, our DualFormer extracts multi-scale global contextual information via a pyramid downsampling module, while Twins only captures global information at a specific scale. Besides, Twins is originally designed for image processing and hence needs elaborate ways to generalize to spatiotemporal domains. 

RegionViT differs from our model in how local tokens interact with global contexts. It generates coarse-grained regional tokens and fine-grained local tokens from an image with different patch sizes, where each regional token is associated with a set of local tokens based on their locations. All regional tokens are first passed through a standard MSA to exchange the information among regions, and then a local self-attention performs MSA where each takes one regional token and corresponding local tokens. In other words, the local token will only interact with the regional token that it belongs to, while each local token in DualFormer \emph{directly} interacts with all multi-scale global contexts.

\section{Experiments}
We evaluate our approach on five popular video datasets. 
For action recognition, we use two versions of Kinetics~\cite{kay2017kinetics}, i.e.,  \textbf{Kinetics-400}/\textbf{Kinetics-600} which contain about 240K/370K training videos and 20k/28k validation videos, and has 400/600 action  classes. For temporal modeling, since the   Something-Something~\cite{goyal2017something} dataset has expired, we test DualFormer on another fine-grained action benchmark, namely~\textbf{Diving-48} \cite{li2018resound} which consists of $\sim$18k videos with 48 diving classes. Finally, we examine transfer learning performance of our method on two smaller datasets, including \textbf{HMDB-51}~\cite{kuehne2011hmdb} and \textbf{UCF-101}~\cite{soomro2012ucf101}.

\subsection{Implementation Details}
Unless otherwise stated, our model receives a clip of 32 frames sampled from the original video using a temporal stride of 2 and spatial resolution of 224$\times$224, yielding  16$\times$56$\times$56 tokens at the first stage. During inference, 4 temporal clips with a center crop (totally 4 space-time views) are exploited to compute accuracy.

\noindent \textbf{Kinetics-400/600.} For both Kinetics datasets, we use AdamW~\cite{kingma2014adam} optimizer with a batch size 64 and a cosine learning rate scheduler to  train DualFormer   for 30 epochs. Following Swin~\cite{liu2021video}, we utilize different initial learning rates for the ImageNet-pretrained backbone (1e-4) and head (1e-3). We also use a linear warm-up for the first 2.5 epochs. To avoid overfitting, we set weight decay to 0.02, 0.02, 0.05 and stochastic depth drop rates~\cite{huang2016deep} to 0.1, 0.2 and 0.3 for the tiny, small and base versions, respectively. Token labeling \cite{jiang2021token} is employed as augmentation to improve DualFormer-T/S. See more details in the Appendix.
 
\noindent \textbf{Diving/HMDB/UCF.} On these three datasets, we adopt AdamW~\cite{kingma2014adam} optimizer to train 16 epochs with one epoch of linear warm-up. The learning rate, batch size, weight decay and stochastic depth drop rate are the same as those for Kinetics. We use the pretrained weights on ImageNet-1K or Kinetics-400 for the model initialization for different settings. 
\begin{table*}[!t]
  \centering
  \scriptsize
  \tabcolsep=0.8mm
    \begin{tabular}{l|c|c|c|c|c|cc|cc}
    \shline
    \multicolumn{1}{l|}{\multirow{2}{*}{Method}} & \multicolumn{1}{c|}{\multirow{2}{*}{Pretrain}} & \multirow{2}{*}{Input} & \multirow{2}{*}{Views} & \multirow{2}{*}{\makecell{Overall \\ FLOPs}} & \multirow{2}{*}{Param} & \multicolumn{2}{c|}{Kinetics-400} & \multicolumn{2}{c}{Kinetics-600} \\
\cline{7-10}     &      &       &       &       &       & Top-1 & Top-5 & Top-1 & Top-5 \\
    \hline
    R(2+1)D~\cite{tran2018r21d} & - & 32 $\times$ 2 & 10 $\times$ 1 & 750 & 61.8  & 72.0  & 90.0  & -  & -  \\
    I3D~\cite{carreira2017i3d} & IN-1K & 32 $\times$ 2  & - & 108 & 25.0  & 72.1  & 90.3  & -  & -  \\
    SlowFast+NL~\cite{feichtenhofer2019slowfast} & - & -  & 10 $\times$ 3 & 7020 & 59.9  & 79.8  & 93.9  & 81.8  & 95.1  \\
    X3D-XL~\cite{feichtenhofer2020x3d} & - & 16 $\times$ 5 & 10 $\times$ 3 & 1452 & 11.0 & 79.1 & 93.9  & 81.9 & 95.5  \\
    X3D-XXL~\cite{feichtenhofer2020x3d} & - & 16  $\times$ 5 & 10 $\times$ 3 & 5823 & 20.3  & 80.4  & 94.6  & - & -  \\
    \color{mygray}{ip-CSN-152~\cite{tran2019video}} & \color{mygray}{IG-65M} & 8 & \color{mygray}{10 $\times$ 3} & \color{mygray}{3270} & \color{mygray}{32.8} & \color{mygray}{82.5} & \color{mygray}{95.3} & - & -  \\
    \hline
    ViT-B-VTN~\cite{neimark2021video} & IN-21K & 250 $\times$ 1 &  1 $\times$ 1 & 4218   & 11.0  & 78.6  & 93.7  & - & -  \\
    TimeSformer-L~\cite{bertasius2021space} & IN-21K & 96 $\times$ 4 & 1 $\times$ 3 & 7140 & 121.4  & 80.7  & 94.7  & 82.2 & 95.5  \\
    MViT-B, 32$\times$3~\cite{fan2021multiscale} & - & 32 $\times$ 3 & 1 $\times$ 5 & 850  & 36.6  & 80.2  & 94.4  & 83.8  & 96.3  \\
    MViT-B, 64$\times$3~\cite{fan2021multiscale} & - & 64 $\times$ 3 & 3 $\times$ 3 & 4095 & 36.6  & 81.2  & 95.1  & - & -  \\
    VidTr-L~\cite{zhang2021vidtr} & IN-21K & 32 $\times$ 2 & 10 $\times$ 3 & 10530  & - & 78.6  & 93.5  & -  & -  \\

    X-ViT (16$\times$)~\cite{bulat2021space} & IN-21K & 16 $\times$ 4 & 1 $\times$ 3 & 850 & -  & 80.2  & 94.7  & 84.5  & 96.3  \\
    ViViT-L/16$\times$2~\cite{arnab2021vivit} & IN-21K & 32 $\times$ 2 & 4 $\times$ 3 & 17352 & 310.8  & 80.6  & 94.7  & 82.5  & 95.6  \\
    \color{mygray}{ViViT-L/16$\times$2~\cite{arnab2021vivit}}  & \color{mygray}{JFT-300M} & 32 $\times$ 2 & \color{mygray}{4 $\times$ 3} & \color{mygray}{17352} & \color{mygray}{310.8} & \color{mygray}{82.8} & \color{mygray}{95.5} & \color{mygray}{84.3}  & \color{mygray}{96.2}  \\

    Swin-T~\cite{liu2021video} & IN-1K & 32 $\times$ 2 & 4 $\times$ 3 & 1056 & 28.2  & 78.8  & 93.6  & -  & -  \\
    Swin-S~\cite{liu2021video} & IN-1K & 32 $\times$ 2 & 4 $\times$ 3 & 1992 & 49.8  & 80.6  & 94.5  & -  & -  \\
    Swin-B~\cite{liu2021video} & IN-1K & 32 $\times$ 2 & 4 $\times$ 3 & 3384 & 88.1  & 80.6  & 94.6  & -  & -  \\
    Swin-B~\cite{liu2021video} & IN-21K & 32 $\times$ 2 & 4 $\times$ 3 & 3384 & 88.1  & 82.7  & \textbf{95.5}  & 84.0  & 96.5 \\
    \hline
    
   \rowcolor{white} DualFormer-T (ours) & IN-1K & 32 $\times$ 2 & 4 $\times$ 1 & 240 & 21.8  & 79.5  & 94.1  & - & - \\
   \rowcolor{white} DualFormer-S (ours) & IN-1K & 32 $\times$ 2 & 4 $\times$ 1 & 636 & 48.9  & 80.6  & 94.9  & - & -  \\
   \rowcolor{white} DualFormer-B (ours) & IN-1K & 32 $\times$ 2 & 4 $\times$ 1 & 1072 & 86.8  & 81.1  & 95.0  & -  & - \\
   \rowcolor{white} DualFormer-B (ours) & IN-21K & 32 $\times$ 2 & 4 $\times$ 1 & 1072 & 86.8 & \textbf{82.9} & \textbf{95.5} & \textbf{85.2} & \textbf{96.6} \\
    \shline
    \end{tabular}%
  \caption{Comparisons with state-of-the-art methods for action recognition on Kinetics-400/600. All models are trained and evaluated on 224$\times$224 spatial resolution. $n \times s$ input indicates we feed $n$ frames to the network sampled every $s$ frames. FLOPs indicates the total floating point operations per second during inference. The magnitudes are Giga ($10^9$) and Mega ($10^6$) for FLOPs and Param, respectively. IN: ImageNet.}
  \label{tab:result1}%
\end{table*}%

\subsection{Comparison to State-of-the-art}
\noindent \textbf{Kinetics-400.} We present the top-1 and top-5 accuracy of CNNs (upper part) and transformer-based methods (lower part) in Table \ref{tab:result1}. Compared to the best CNN-based method X3D-XXL \cite{xie2018s3d}, DualFormer-S achieves slightly higher accuracy while using \textbf{9.2$\times$} fewer FLOPs. Compared to transformers (MViT-B,32$\times$3 \cite{fan2021multiscale} and X-ViT \cite{bulat2021space}), DualFormer-S with similar computations brings $\sim$0.4\% gain on the top-1 accuracy. In contrast to Swin-T \cite{liu2021video}, DualFormer-T outperforms it by 0.7\% on top-1 and 0.5\% on top-5 score with \textbf{4.4$\times$} fewer computational costs. We also witness 1.8\% improvement on the top-1 accuracy when using ImageNet-21K to pretrain DualFormer-B compared to ImageNet-1K. With ImageNet-21K pretraining, DualFormer-B achieves the state-of-the-art results on both metrics while being dramatically faster than two recent transformer backbones: \textbf{16.2$\times$} faster than ViViT-L \cite{arnab2021vivit} and \textbf{3.2$\times$} faster than Swin-B \cite{liu2021video}. See more details on accuracy vs. speed in Fig. \ref{fig_acc_flops}.

\noindent \textbf{Kinetics-600.} 
As shown in Table \ref{tab:result1}, the results on Kinetics-600 are similar to those on Kinetics-400. DualFormer-B achieves the highest accuracy among these models. In particular, DualFormer-B brings \textbf{1.2\%} gains on top-1 score and runs 3.2$\times$ faster than Swin-B. Compared to ViViT-L which is pretrained on a large-scale and private dataset JFM-300M, although our DualFormer-B is pretrained on a much smaller dataset (ImageNet-21K), it yields \textbf{0.9}\% higher top-1 accuracy and requires \textbf{16.2$\times$} fewer FLOPs.

\noindent \textbf{Diving-48.} Here we test our model on a temporally-heavy dataset. Due to a recently reported label issue of Diving-48, we only compare our model with SlowFast~\cite{feichtenhofer2019slowfast} and TimeSformer~\cite{bertasius2021space}. 
From Table \ref{tab:result2}, we observe that our DualFormer obtains a maximum \textbf{81.8\%} top-1 score on Diving-48, significantly surpassing SlowFast. For TimeSformer-L which has \textbf{3.7$\times$} FLOPs and receives 96 frames as input, our method still yields \textbf{0.8\%} higher accuracy while using only 32 frames as input. These results verify the strong power of our model in temporal modeling.

\noindent \textbf{HMDB-51 and UCF-101.} Lastly, we examine the transfer learning ability of our DualFormer over the split 1 of HMDB-51 and UCF-101. Table \ref{tab:result2} reports the top-1 accuracy. With ImageNet-1K pretrained weights as initialization, our tiny version achieves comparable performance to VidTr-M \cite{zhang2021vidtr} while using \textbf{192$\times$} fewer FLOPs (see DualFormer-T* in Table \ref{tab:result2}). When pretrained on Kinetics-400, DualFormer-S with 12 testing views can outperform VidTr-L by a large accuracy  margin \textbf{2\%}/0.8\%   on HMDB and UCF while using only \textbf{18\%}  FLOPs of VidTr-L. This reveals the generalization potential of our model on small datasets.

\begin{table}[!t]
  \centering
  \scriptsize
  \tabcolsep=2.6mm
    \begin{tabular}{l|c|c|c|c|c|c}
    \shline
    Method & Input & Views & \multicolumn{1}{l|}{FLOPs} & \multicolumn{1}{l|}{DIVE} & \multicolumn{1}{l|}{HMDB} & \multicolumn{1}{l}{UCF} \\
    \hline
    I3D~\cite{carreira2017i3d}  & 64$\times$1  & - & - &  - & 74.3  & 95.1 \\
    TSM~\cite{lin2019tsm} & 8 & - & -  & - & 70.7  & 94.5  \\
    TeiNet~\cite{liu2020teinet} & 16 & - & - & - & 73.3  & 96.7 \\
    SlowFast~\cite{feichtenhofer2019slowfast} & 16$\times$8  & - & - & 77.6 & - & -  \\
    \hline
    VidTr-M~\cite{zhang2021vidtr} & 16$\times$4  & 10$\times$3  & 5370  & - & 74.4  & 96.6 \\
    VidTr-L~\cite{zhang2021vidtr} & 32$\times$4  & 10$\times$3  & 10530 & - & 74.4  & 96.7 \\
    TimeSformer~\cite{bertasius2021space} & 8$\times$4   & 1$\times$3   & 590 & 75.0 & - & -  \\
    TimeSformer-L~\cite{bertasius2021space} & 96$\times$4  & 1$\times$3   & 7140 & 81.0 & - & -  \\
    \hline
    \rowcolor{white} DualFormer-T* & 16$\times$4  & 4$\times$1  & 28  & 75.4 & 74.6   & 96.3 \\
    \rowcolor{white} DualFormer-T & 16$\times$4  & 4$\times$1   & 28   & 75.9 & 75.0 & 96.6 \\
    \rowcolor{white} DualFormer-S & 32$\times$4  & 4$\times$1   & 636  & 81.2  & 76.2   & 97.4 \\
    \rowcolor{white} DualFormer-S & 32$\times$4  & 4$\times$3   & 1908 & \textbf{81.8} & \textbf{76.4} & \textbf{97.5} \\
    \shline
    \end{tabular}%
  \caption{Results on HMDB-51, UCF-101 and Diving-48 (DIVE).  Baseline results are from \protect\cite{bertasius2021space,zhang2021vidtr}.
  We pretrain our models on Kinetics-400 and finetune them on these datasets, only except for DualFormer-T* which is pretrained on ImageNet-1K.}
  \label{tab:result2}%
\end{table}

\begin{minipage}[!t]{\linewidth}
    \hspace{-1.5em}
    \begin{minipage}[!t]{0.55\linewidth}
    \centering
		\renewcommand{\arraystretch}{1.35}
    \scriptsize
    \begin{tabular}{l|c|c|c|c}
    \shline
    Variants & FLOPs & Param & Top-1 & Top-5 \\
    \hline
    (LL, LL, LL, LL) & 244   & 21.7  & 78.4  & 93.3  \\
    (GG, GG, GG, GG) & 228   & 21.8  & 77.6  & 93.2  \\
    (LL, LL, LG, LG) & 236   & 21.7  & 78.8  & 93.5  \\
    (LG, LG, LL, LL) & 244   & 21.8  & 79.3  & 94.0  \\
    \hline
    (LG$_1$, LG$_1$, LG$_1$, LG$_1$) & 224   & 21.8  & 78.4  & 93.4  \\
    (LG$_2$, LG$_2$, LG$_2$, LG$_2$) & 232   & 21.8  & 79.3  & 93.9  \\
    \rowcolor{background_gray} (LG, LG, LG, LG) & 240   & 21.8  & 79.5  & 94.1  \\
    \shline
    \end{tabular}%
    \makeatletter\def\@captype{table}\makeatother\caption{Experimental results of different combinations of LW-MSA (L) and GP-MSA (G) with DualFormer-T on Kinetics-400. G$_1$ and G$_2$ denote GP-MSA with only one pyramid scale (4,4,4) and (8,7,7), respectively. The gray row indicates our default setting.}\label{tab:dualmsa}%
    \end{minipage}
    \hspace{0.2em}
    \begin{minipage}[!t]{0.41\linewidth}
    \centering
    \includegraphics[width=.95\linewidth]{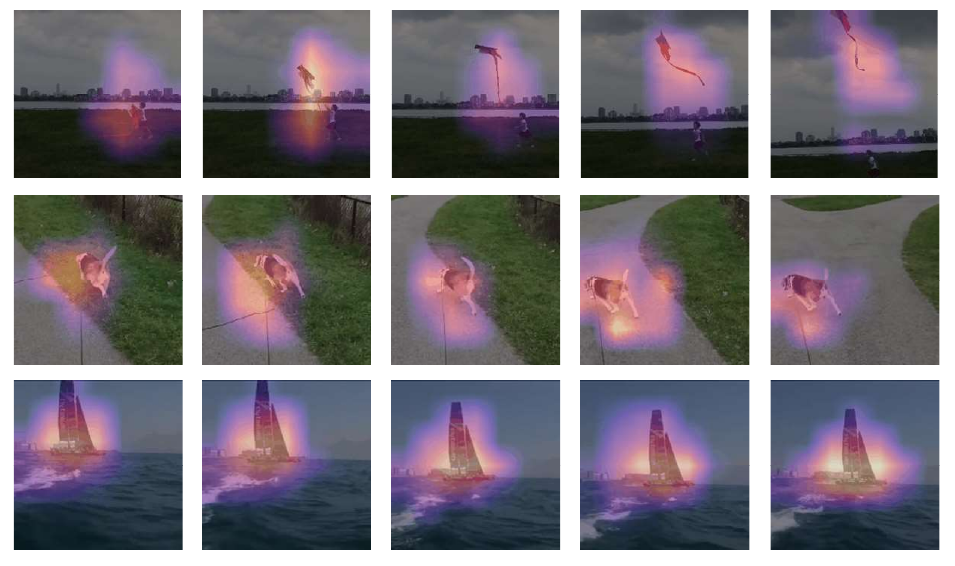}
  \makeatletter\def\@captype{figure}\makeatother\caption{Visualization of attention maps at the last layer generated by Grad-CAM \protect~\cite{selvaraju2017grad} on Kinetics-400. Our model successfully learns to focus on the relevant parts in the video clip. Upper: flying kites. Middle: walking dogs. Below: sailing.}
  \label{fig_vis}
    \end{minipage}
\end{minipage}

\subsection{Ablation Study}\label{para:ablation}
\noindent \textbf{Effect of LW-MSA \& GP-MSA.} 
To study the effect of the dual-level MSA, we test different combinations of LW/GP-MSA to implement DualFormer-T. (LG, LG, LG, LG) denotes our default configuration, namely the one in Figure \ref{tab:dualmsa},  where each block sequentially performs LW-MSA and GP-MSA. For the four variants at the upper part of Table \ref{tab:dualmsa}, LL and GG mean that the blocks at that stage only contain two LW-MSAs and two GP-MSAs, respectively. For example, (LL, LL, LG, LG) means using blocks with two LW-MSAs at the first two stages and using a combination of LW-MSA and GP-MSA at the last two stages.

For a fair comparison, we slightly tune the hyperparameter to ensure their FLOPs and parameters to be similar. We report the accuracy of these variants on Kinetics-400 in the upper part of Table \ref{tab:dualmsa}.
Among these variants, (GG, GG, GG, GG) performs the worst since the local context information is very important to a patch. The model with only LW-MSA degrades by 1.1\% top-1 score (79.5\%$\rightarrow$78.4\%) due to a limited receptive field at every stage. By integrating GP-MSA to increase the receptive field, both (LL, LL, LG, LG) and (LG, LG, LL, LL) achieve better performance than the variants with only local or global modules. In particular, adding GP-MSA to the early stages benefits more than late stages, revealing the importance of GP-MSA to complement the early stages. Moreover, we evaluate the two pyramid scales in GP-MSA and report their results in the lower part of Table \ref{tab:dualmsa}. Compared to our default setting, we can find a clear accuracy drop by removing either the (4, 4, 4) or (8, 7, 7) scale. In addition, some examples of attention visualization are shown in Figure \ref{fig_vis}.

\begin{figure}[!t]
  \centering
  \includegraphics[width=0.7\textwidth]{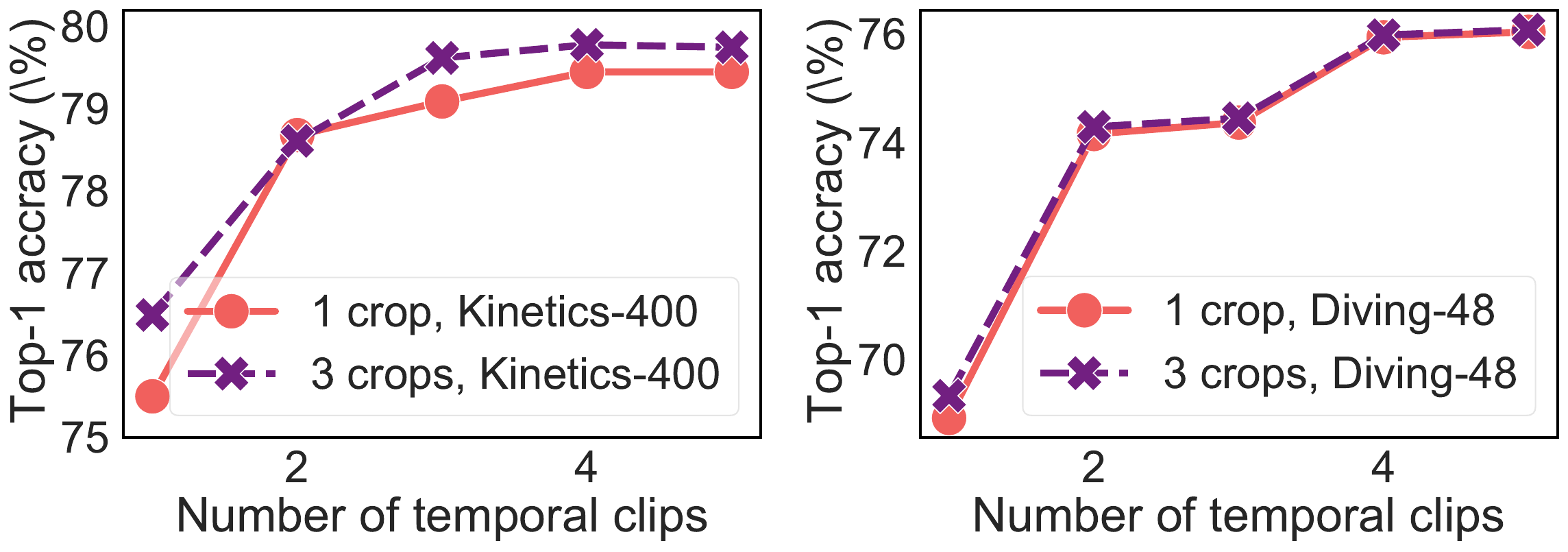}
  \caption{Effect of space-time views on Kinetics-400 (left) and on Diving-48 (right).}
  \label{fig_views}
\end{figure}

\noindent \textbf{Effect of testing views.} Previous methods employ multiple space-time views to boost performance during inference, e.g., 10$\times$3 views in VidTr-L and 4$\times$3 views in Swin. We investigate how the number of testing views affects the accuracy of DualFormer-T on Kinetics-400 and Diving-48. From Figure \ref{fig_views}, one can find that increasing the number of temporal clips can bring significant improvement on both datasets, while using more spatial crops does not always help. For example, using three spatial crops slightly outperforms the 1-crop counterpart on Diving-48. As the inference FLOPs is proportional to the space-time views, to trade off the computational cost  and accuracy, our method uses a testing strategy of four temporal clips with a spatial crop (totally four) during the inference phase.

\begin{table}[!b]
\scriptsize
\begin{floatrow}
\capbtabbox{
 \tabcolsep=2mm
 \renewcommand{\arraystretch}{0.85}
  \scriptsize
    \begin{tabular}{l|c|c|c|c}
    \shline
    Input & Window Size & FLOPs & Top-1 & Top-5 \\
    \hline
    16$\times$4  & 4$\times$7$\times$7 & 104 & 78.0  & 93.2  \\
    16$\times$4  & 8$\times$7$\times$7 & 112 & 78.4  & 93.3  \\
    \hline
    32$\times$2  & 4$\times$7$\times$7 & 224 & 79.1  & 93.9  \\
    \rowcolor{background_gray} 32$\times$2  & 8$\times$7$\times$7 & 240 & 79.5 & 94.1  \\
    32$\times$2  & 16$\times$7$\times$7 & 272 & \textbf{79.7}  & 94.4  \\
    32$\times$2  & 8$\times$14$\times$14 & 324 & \textbf{79.7}  & \textbf{94.5}  \\
    \shline
    \end{tabular}%
}{
 \caption{Effect of window size of LW-MSA with DualFormer-T on Kinetic-400. The gray row indicates the default configuration.}
  \label{tab:ws}
}
\capbtabbox{
 \tabcolsep=1mm
		\renewcommand{\arraystretch}{1.5}
 \begin{tabular}{l|c|c|c}
 \shline
 Method & FLOPs & Param & Top-1 \\
 \hline
 AvgPool & 59 & 21.8 & 78.7 \\
 Conv & 61 & 27.6 & \textbf{79.5} \\
 \rowcolor{background_gray} DWConv & 60 & 21.8 & \textbf{79.5} \\
 \shline
 \end{tabular}
}{
 \caption{Results of pyramid downsampling functions based on DualFormer-T on Kinetics-400.}
 \label{tab:pd_func}
}
\end{floatrow}
\end{table}

\noindent \textbf{Effect of window size in LW-MSA.}  Window size is a crucial hyperparameter in LW-MSA. Hence, we test different window sizes to investigate their effect on model performance. As shown in Table \ref{tab:ws}, a larger window size in both temporal and spatial dimensions brings consistent gains in accuracy due to the increase of local receptive field, but also induces heavier computation. For an accuracy-speed balance, we choose $(8,7,7)$ as our default setting. From this table, we also observe that reducing the number of input frames (e.g., 32$\rightarrow$16) can dramatically improve efficiency but inevitably degrades the top-1 accuracy by $\sim$1\%. 

\noindent \textbf{Effect of pyramid downsampling function.} There are several alternative functions to generate global priors in GP-MSA, such as average pooling (AvgPool) and standard convolution (Conv). Here, we replace the depth-wise convolution (DWConv) with them on Kinetics-400 to investigate their effect. As reported in Table \ref{tab:pd_func}, our DWConv achieves comparable performance to Conv while using much fewer parameters. Our implementation also outperforms AvgPool by 0.8\% on the top-1 score with similar computation costs.

\noindent \textbf{Do we need PEG?} As depicted in Table \ref{tab:peg}, DualFormer without PEG suffers from a clear drop on the top-1 accuracy (79.5\%$\rightarrow$78.9\%), which indicates the necessity of integrating position information in MSA. We further compare our DWConv-based PEG with an absolute position encoding (i.e., TimeSformer) and a relative bias-based method in Swin. As a result, our solution achieves 0.3\% and 0.2\% higher top-1 score than the absolute and relative method, respectively.

\noindent \textbf{Effect of Temporal Pooling Rate.}
Our method follows~\cite{fan2021multiscale,liu2021video} to utilize a multi-scale hierarchy. Such hierarchy is achieved by the patch merging layer at the beginning of the last three stages, where we downsample the spatial size of feature map by 2$\times$ and keep the original temporal resolution. Here, we discuss the effect of temporal pooling at the last three stages. According to the results in Table \ref{tab:tp}, even though such temporal pooling can further reduce the computational cost, it also leads to a decrease in the overall accuracy.
\begin{table}[!t]
\scriptsize
\begin{floatrow}
\capbtabbox{
 \tabcolsep=3mm
 \renewcommand{\arraystretch}{1.18}
 \begin{tabular}{l|c}
 \shline
 Method & Top-1 \\
 \hline
 w.o PEG & 78.9 \\
 Absolute~\cite{bertasius2021space} & 79.2 \\
 Relative~\cite{liu2021video} & 79.3   \\
 \rowcolor{background_gray} DWConv & \textbf{79.5} \\
 \shline
 \end{tabular}
}{
 \caption{Effect of PEGs to DualFormer-T on the Kinetics-400 dataset.}
 \label{tab:peg}
}
\capbtabbox{
 \tabcolsep=2mm
  \scriptsize
    \begin{tabular}{l|c|c|c|c|c}
    \shline
    Rate & Patch Size & FLOPs & Param & Top-1 & Top-5 \\
    \hline
    $1,1,1$ & $(4,4,4)$ & 112   & 21.8  & 78.5  & 93.3  \\
    $2,1,1$ & $(2,4,4)$ & 136   & 21.8  & 78.7  & 93.5  \\
    $1,2,1$ & $(2,4,4)$ & 152   & 21.9  & 78.8  & 93.5  \\
    $1,1,2$ & $(2,4,4)$ & 216   & 22.3  & 79.2  & 93.9  \\
    \rowcolor{background_gray} $1,1,1$ & $(2,4,4)$ & 240   & 21.8  & \textbf{79.5}  & \textbf{94.1}  \\
    \shline
    \end{tabular}%
}{
\caption{Effect of temporal pooling in DualFormer-T on Kinetic-400. $(i, j, k)$ means reducing the temporal resolution $i$, $j$, $k$ times at the last 3 stages, respectively.}
\label{tab:tp}
}
\end{floatrow}
\end{table}

\section{Conclusion}
In this paper, we develop a transformer-based architecture with local-global attention stratification for efficient video recognition. Empirical study demonstrates that the proposed method achieves a better accuracy-speed trade-off on five popular video recognition datasets. In the future, we plan to remove the strong dependency on pretrained models and design a useful strategy to train our model from scratch. Another direction is to explore the use of our model in other applications, such as video segmentation and prediction.

\section*{Acknowledgement}
We thank Quanhong Fu at Sea AI Lab for the help to improve the paper writing.  This research is supported by Singapore Ministry of Education Academic Research Fund Tier 1 under MOE's official grant number T1 251RES2029.

\clearpage

\bibliographystyle{splncs04}
\bibliography{egbib}

\begin{thebibliography}{10}
\providecommand{\url}[1]{\texttt{#1}}
\providecommand{\urlprefix}{URL }
\providecommand{\doi}[1]{https://doi.org/#1}

\bibitem{arnab2021vivit}
Arnab, A., Dehghani, M., Heigold, G., Sun, C., Lu{\v{c}}i{\'c}, M., Schmid, C.:
  Vivit: A video vision transformer. arXiv preprint arXiv:2103.15691  (2021)

\bibitem{ba2016layer}
Ba, J.L., Kiros, J.R., Hinton, G.E.: Layer normalization. arXiv preprint
  arXiv:1607.06450  (2016)

\bibitem{bertasius2021space}
Bertasius, G., Wang, H., Torresani, L.: Is space-time attention all you need
  for video understanding? arXiv preprint arXiv:2102.05095  (2021)

\bibitem{bulat2021space}
Bulat, A., Perez-Rua, J.M., Sudhakaran, S., Martinez, B., Tzimiropoulos, G.:
  Space-time mixing attention for video transformer. arXiv preprint
  arXiv:2106.05968  (2021)

\bibitem{carreira2017i3d}
Carreira, J., Zisserman, A.: Quo vadis, action recognition? a new model and the
  kinetics dataset. In: proceedings of the IEEE Conference on Computer Vision
  and Pattern Recognition. pp. 6299--6308 (2017)

\bibitem{chen2021regionvit}
Chen, C.F., Panda, R., Fan, Q.: Regionvit: Regional-to-local attention for
  vision transformers. arXiv preprint arXiv:2106.02689  (2021)

\bibitem{chollet2017xception}
Chollet, F.: Xception: Deep learning with depthwise separable convolutions. In:
  Proceedings of the IEEE conference on computer vision and pattern
  recognition. pp. 1251--1258 (2017)

\bibitem{chu2021Twins}
Chu, X., Tian, Z., Wang, Y., Zhang, B., Ren, H., Wei, X., Xia, H., Shen, C.:
  Twins: Revisiting the design of spatial attention in vision transformers. In:
  NeurIPS 2021 (2021)

\bibitem{chu2021we}
Chu, X., Tian, Z., Zhang, B., Wang, X., Wei, X., Xia, H., Shen, C.: Conditional
  positional encodings for vision transformers. arXiv preprint arXiv:2102.10882
   (2021)

\bibitem{dosovitskiy2020image}
Dosovitskiy, A., Beyer, L., Kolesnikov, A., Weissenborn, D., Zhai, X.,
  Unterthiner, T., Dehghani, M., Minderer, M., Heigold, G., Gelly, S., et~al.:
  An image is worth 16x16 words: Transformers for image recognition at scale.
  In: International Conference on Learning Representations (2020)

\bibitem{fan2021multiscale}
Fan, H., Xiong, B., Mangalam, K., Li, Y., Yan, Z., Malik, J., Feichtenhofer,
  C.: Multiscale vision transformers. arXiv preprint arXiv:2104.11227  (2021)

\bibitem{feichtenhofer2020x3d}
Feichtenhofer, C.: X3d: Expanding architectures for efficient video
  recognition. In: Proceedings of the IEEE/CVF Conference on Computer Vision
  and Pattern Recognition. pp. 203--213 (2020)

\bibitem{feichtenhofer2019slowfast}
Feichtenhofer, C., Fan, H., Malik, J., He, K.: Slowfast networks for video
  recognition. In: Proceedings of the IEEE/CVF international conference on
  computer vision. pp. 6202--6211 (2019)

\bibitem{goyal2017something}
Goyal, R., Ebrahimi~Kahou, S., Michalski, V., Materzynska, J., Westphal, S.,
  Kim, H., Haenel, V., Fruend, I., Yianilos, P., Mueller-Freitag, M., et~al.:
  The" something something" video database for learning and evaluating visual
  common sense. In: Proceedings of the IEEE international conference on
  computer vision. pp. 5842--5850 (2017)

\bibitem{hara2017learning}
Hara, K., Kataoka, H., Satoh, Y.: Learning spatio-temporal features with 3d
  residual networks for action recognition. In: Proceedings of the IEEE
  International Conference on Computer Vision Workshops. pp. 3154--3160 (2017)

\bibitem{hara2018can}
Hara, K., Kataoka, H., Satoh, Y.: Can spatiotemporal 3d cnns retrace the
  history of 2d cnns and imagenet? In: Proceedings of the IEEE conference on
  Computer Vision and Pattern Recognition. pp. 6546--6555 (2018)

\bibitem{he2016deep}
He, K., Zhang, X., Ren, S., Sun, J.: Deep residual learning for image
  recognition. In: Proceedings of the IEEE conference on computer vision and
  pattern recognition. pp. 770--778 (2016)

\bibitem{hongeng2004video}
Hongeng, S., Nevatia, R., Bremond, F.: Video-based event recognition: activity
  representation and probabilistic recognition methods. Computer Vision and
  Image Understanding  \textbf{96}(2),  129--162 (2004)

\bibitem{huang2016deep}
Huang, G., Sun, Y., Liu, Z., Sedra, D., Weinberger, K.Q.: Deep networks with
  stochastic depth. In: European conference on computer vision. pp. 646--661.
  Springer (2016)

\bibitem{islam2020much}
Islam, M.A., Jia, S., Bruce, N.D.: How much position information do
  convolutional neural networks encode? arXiv preprint arXiv:2001.08248  (2020)

\bibitem{ji20123d}
Ji, S., Xu, W., Yang, M., Yu, K.: 3d convolutional neural networks for human
  action recognition. IEEE transactions on pattern analysis and machine
  intelligence  \textbf{35}(1),  221--231 (2012)

\bibitem{jiang2019stm}
Jiang, B., Wang, M., Gan, W., Wu, W., Yan, J.: Stm: Spatiotemporal and motion
  encoding for action recognition. In: Proceedings of the IEEE/CVF
  International Conference on Computer Vision. pp. 2000--2009 (2019)

\bibitem{jiang2021token}
Jiang, Z., Hou, Q., Yuan, L., Zhou, D., Jin, X., Wang, A., Feng, J.: Token
  labeling: Training a 85.5\% top-1 accuracy vision transformer with 56m
  parameters on imagenet. arXiv preprint arXiv:2104.10858  (2021)

\bibitem{karpathy2014large}
Karpathy, A., Toderici, G., Shetty, S., Leung, T., Sukthankar, R., Fei-Fei, L.:
  Large-scale video classification with convolutional neural networks. In:
  Proceedings of the IEEE conference on Computer Vision and Pattern
  Recognition. pp. 1725--1732 (2014)

\bibitem{kay2017kinetics}
Kay, W., Carreira, J., Simonyan, K., Zhang, B., Hillier, C., Vijayanarasimhan,
  S., Viola, F., Green, T., Back, T., Natsev, P., et~al.: The kinetics human
  action video dataset. arXiv preprint arXiv:1705.06950  (2017)

\bibitem{kingma2014adam}
Kingma, D.P., Ba, J.: Adam: A method for stochastic optimization. arXiv
  preprint arXiv:1412.6980  (2014)

\bibitem{krizhevsky2012imagenet}
Krizhevsky, A., Sutskever, I., Hinton, G.E.: Imagenet classification with deep
  convolutional neural networks. Advances in neural information processing
  systems  \textbf{25},  1097--1105 (2012)

\bibitem{kuehne2011hmdb}
Kuehne, H., Jhuang, H., Garrote, E., Poggio, T., Serre, T.: Hmdb: a large video
  database for human motion recognition. In: 2011 International conference on
  computer vision. pp. 2556--2563. IEEE (2011)

\bibitem{li2021uniformer}
Li, K., Wang, Y., Peng, G., Song, G., Liu, Y., Li, H., Qiao, Y.: Uniformer:
  Unified transformer for efficient spatial-temporal representation learning.
  In: International Conference on Learning Representations (2021)

\bibitem{li2018resound}
Li, Y., Li, Y., Vasconcelos, N.: Resound: Towards action recognition without
  representation bias. In: Proceedings of the European Conference on Computer
  Vision (ECCV). pp. 513--528 (2018)

\bibitem{li2021survey}
Li, Z., Liu, F., Yang, W., Peng, S., Zhou, J.: A survey of convolutional neural
  networks: analysis, applications, and prospects. IEEE Transactions on Neural
  Networks and Learning Systems  (2021)

\bibitem{lin2019tsm}
Lin, J., Gan, C., Han, S.: Tsm: Temporal shift module for efficient video
  understanding. In: Proceedings of the IEEE/CVF International Conference on
  Computer Vision. pp. 7083--7093 (2019)

\bibitem{lin2014network}
Lin, M., Chen, Q., Yan, S.: Network in network. In: Bengio, Y., LeCun, Y.
  (eds.) 2nd International Conference on Learning Representations, {ICLR} 2014,
  Banff, AB, Canada, April 14-16, 2014, Conference Track Proceedings (2014)

\bibitem{liu2021Swin}
Liu, Z., Lin, Y., Cao, Y., Hu, H., Wei, Y., Zhang, Z., Lin, S., Guo, B.: Swin
  transformer: Hierarchical vision transformer using shifted windows.
  International Conference on Computer Vision (ICCV)  (2021)

\bibitem{liu2021video}
Liu, Z., Ning, J., Cao, Y., Wei, Y., Zhang, Z., Lin, S., Hu, H.: Video swin
  transformer. arXiv preprint arXiv:2106.13230  (2021)

\bibitem{liu2020teinet}
Liu, Z., Luo, D., Wang, Y., Wang, L., Tai, Y., Wang, C., Li, J., Huang, F., Lu,
  T.: Teinet: Towards an efficient architecture for video recognition. In:
  Proceedings of the AAAI Conference on Artificial Intelligence. vol.~34, pp.
  11669--11676 (2020)

\bibitem{liu2021tam}
Liu, Z., Wang, L., Wu, W., Qian, C., Lu, T.: Tam: Temporal adaptive module for
  video recognition. In: Proceedings of the IEEE/CVF International Conference
  on Computer Vision. pp. 13708--13718 (2021)

\bibitem{neimark2021video}
Neimark, D., Bar, O., Zohar, M., Asselmann, D.: Video transformer network.
  arXiv preprint arXiv:2102.00719  (2021)

\bibitem{patrick2021keeping}
Patrick, M., Campbell, D., Asano, Y.M., Metze, I.M.F., Feichtenhofer, C.,
  Vedaldi, A., Henriques, J., et~al.: Keeping your eye on the ball: Trajectory
  attention in video transformers. arXiv preprint arXiv:2106.05392  (2021)

\bibitem{qian2021spatiotemporal}
Qian, R., Meng, T., Gong, B., Yang, M.H., Wang, H., Belongie, S., Cui, Y.:
  Spatiotemporal contrastive video representation learning. In: Proceedings of
  the IEEE/CVF Conference on Computer Vision and Pattern Recognition. pp.
  6964--6974 (2021)

\bibitem{qiu2017p3d}
Qiu, Z., Yao, T., Mei, T.: Learning spatio-temporal representation with
  pseudo-3d residual networks. In: proceedings of the IEEE International
  Conference on Computer Vision. pp. 5533--5541 (2017)

\bibitem{ranftl2021vision}
Ranftl, R., Bochkovskiy, A., Koltun, V.: Vision transformers for dense
  prediction. In: Proceedings of the IEEE/CVF International Conference on
  Computer Vision. pp. 12179--12188 (2021)

\bibitem{selvaraju2017grad}
Selvaraju, R.R., Cogswell, M., Das, A., Vedantam, R., Parikh, D., Batra, D.:
  Grad-cam: Visual explanations from deep networks via gradient-based
  localization. In: Proceedings of the IEEE international conference on
  computer vision. pp. 618--626 (2017)

\bibitem{simonyan2014very}
Simonyan, K., Zisserman, A.: Very deep convolutional networks for large-scale
  image recognition. arXiv preprint arXiv:1409.1556  (2014)

\bibitem{soomro2012ucf101}
Soomro, K., Zamir, A.R., Shah, M.: Ucf101: A dataset of 101 human actions
  classes from videos in the wild. arXiv preprint arXiv:1212.0402  (2012)

\bibitem{tobler1970computer}
Tobler, W.R.: A computer movie simulating urban growth in the detroit region.
  Economic geography  \textbf{46}(sup1),  234--240 (1970)

\bibitem{touvron2021training}
Touvron, H., Cord, M., Douze, M., Massa, F., Sablayrolles, A., J{\'e}gou, H.:
  Training data-efficient image transformers \& distillation through attention.
  In: International Conference on Machine Learning. pp. 10347--10357. PMLR
  (2021)

\bibitem{du2015c3d}
Tran, D., Bourdev, L., Fergus, R., Torresani, L., Paluri, M.: Learning
  spatiotemporal features with 3d convolutional networks. In: Proceedings of
  the IEEE international conference on computer vision. pp. 4489--4497 (2015)

\bibitem{tran2019video}
Tran, D., Wang, H., Torresani, L., Feiszli, M.: Video classification with
  channel-separated convolutional networks. In: Proceedings of the IEEE/CVF
  International Conference on Computer Vision. pp. 5552--5561 (2019)

\bibitem{tran2018r21d}
Tran, D., Wang, H., Torresani, L., Ray, J., LeCun, Y., Paluri, M.: A closer
  look at spatiotemporal convolutions for action recognition. In: Proceedings
  of the IEEE conference on Computer Vision and Pattern Recognition. pp.
  6450--6459 (2018)

\bibitem{vaswani2017attention}
Vaswani, A., Shazeer, N., Parmar, N., Uszkoreit, J., Jones, L., Gomez, A.N.,
  Kaiser, {\L}., Polosukhin, I.: Attention is all you need. In: Advances in
  neural information processing systems. pp. 5998--6008 (2017)

\bibitem{wang2021max}
Wang, H., Zhu, Y., Adam, H., Yuille, A., Chen, L.C.: Max-deeplab: End-to-end
  panoptic segmentation with mask transformers. In: Proceedings of the IEEE/CVF
  Conference on Computer Vision and Pattern Recognition. pp. 5463--5474 (2021)

\bibitem{wang2018temporal}
Wang, L., Xiong, Y., Wang, Z., Qiao, Y., Lin, D., Tang, X., Van~Gool, L.:
  Temporal segment networks for action recognition in videos. IEEE transactions
  on pattern analysis and machine intelligence  \textbf{41}(11),  2740--2755
  (2018)

\bibitem{wang2021pyramid}
Wang, W., Xie, E., Li, X., Fan, D.P., Song, K., Liang, D., Lu, T., Luo, P.,
  Shao, L.: Pyramid vision transformer: A versatile backbone for dense
  prediction without convolutions. arXiv preprint arXiv:2102.12122  (2021)

\bibitem{wang2021end}
Wang, Y., Xu, Z., Wang, X., Shen, C., Cheng, B., Shen, H., Xia, H.: End-to-end
  video instance segmentation with transformers. In: Proceedings of the
  IEEE/CVF Conference on Computer Vision and Pattern Recognition. pp.
  8741--8750 (2021)

\bibitem{wei2021masked}
Wei, C., Fan, H., Xie, S., Wu, C.Y., Yuille, A., Feichtenhofer, C.: Masked
  feature prediction for self-supervised visual pre-training. arXiv preprint
  arXiv:2112.09133  (2021)

\bibitem{xie2018s3d}
Xie, S., Sun, C., Huang, J., Tu, Z., Murphy, K.: Rethinking spatiotemporal
  feature learning: Speed-accuracy trade-offs in video classification. In:
  Proceedings of the European conference on computer vision (ECCV). pp.
  305--321 (2018)

\bibitem{xu2017r}
Xu, H., Das, A., Saenko, K.: R-c3d: Region convolutional 3d network for
  temporal activity detection. In: Proceedings of the IEEE international
  conference on computer vision. pp. 5783--5792 (2017)

\bibitem{yang2021focal}
Yang, J., Li, C., Zhang, P., Dai, X., Xiao, B., Yuan, L., Gao, J.: Focal
  self-attention for local-global interactions in vision transformers. arXiv
  preprint arXiv:2107.00641  (2021)

\bibitem{zha2021shifted}
Zha, X., Zhu, W., Lv, T., Yang, S., Liu, J.: Shifted chunk transformer for
  spatio-temporal representational learning. arXiv preprint arXiv:2108.11575
  (2021)

\bibitem{zhang2021vidtr}
Zhang, Y., Li, X., Liu, C., Shuai, B., Zhu, Y., Brattoli, B., Chen, H., Marsic,
  I., Tighe, J.: Vidtr: Video transformer without convolutions. In: Proceedings
  of the IEEE/CVF International Conference on Computer Vision. pp. 13577--13587
  (2021)

\bibitem{zhu2020deformable}
Zhu, X., Su, W., Lu, L., Li, B., Wang, X., Dai, J.: Deformable detr: Deformable
  transformers for end-to-end object detection. arXiv preprint arXiv:2010.04159
   (2020)

\end{thebibliography}
\end{document}